\documentclass[a4paper,fleqn]{cas-sc}

\usepackage{tabularray}
\usepackage[numbers]{natbib}
\def\tsc#1{\csdef{#1}{\textsc{\lowercase{#1}}\xspace}}
\tsc{WGM}
\tsc{QE}
\begin{document}
\let\WriteBookmarks\relax
\def\floatpagepagefraction{1}
\def\textpagefraction{.001}
\let\printorcid\relax % 可去掉页面下方的ORCID(s)

\shorttitle{R2GenGPT}    
% \shorttitle{Leveraging social media news to predict stock index movement using RNN-boost}   

\shortauthors{Zhanyu Wang et al.} 
% \shortauthors{V. {{\=A}}nand Rawat et al.}

\title[mode = title]{R2GenGPT: Radiology Report Generation with Frozen LLMs}

\author[1]{Zhanyu Wang}[style=chinese, orcid=0000-0000-0000-0000] 
% \cormark[1] 
\ead{zhanyu.wang@sydney.edu.au}
\ead[url]{https://wang-zhanyu.github.io/}
\credit{Conceptualization of this study, Methodology, Software}

\author[2]{Lingqiao Liu}[style=chinese, orcid=0000-0000-0000-0000] 
\ead{lingqiao.liu@adelaide.edu.au}
\ead[URL]{https://lingqiao-adelaide.github.io/lingqiaoliu.github.io/}

\author[3]{Lei Wang}[style=chinese, orcid=0000-0000-0000-0000]
\ead{leiw@uow.edu.au}
\ead[URL]{https://sites.google.com/view/lei-hs-wang}

\author[1]{Luping Zhou}[style=chinese, orcid=0000-0000-0000-0000]
\cormark[1] 
\ead{luping.zhou@sydney.edu.au}
\ead[URL]{https://sites.google.com/view/lupingzhou}
\credit{Data curation, Writing - Original draft preparation}

\address[1]{University of Sydney, New South Wales 2006, Australia}
\address[2]{University of Adelaide, South Australia 5005, Australia}
\address[3]{University of Wollongong, New South Wales 2522, Australia}

\cortext[1]{Corresponding author} 

\begin{abstract}
Large Language Models (LLMs) have consistently showcased remarkable generalization capabilities when applied to various language tasks. Nonetheless, harnessing the full potential of LLMs for Radiology Report Generation (R2Gen) still presents a challenge, stemming from the inherent disparity in modality between LLMs and the R2Gen task. To bridge this gap effectively, we propose R2GenGPT, which is a novel solution that aligns visual features with the word embedding space of LLMs using an efficient visual alignment module. This innovative approach empowers the previously static LLM to seamlessly integrate and process image information, marking a step forward in optimizing R2Gen performance.
R2GenGPT offers the following benefits. First, it attains state-of-the-art (SOTA) performance by training only the lightweight visual alignment module while freezing all the parameters of LLM. Second, it exhibits high training efficiency, as it requires the training of an exceptionally minimal number of parameters while achieving rapid convergence. By employing delta tuning, our model only trains 5M parameters (which constitute just 0.07\% of the total parameter count) to achieve performance close to the SOTA levels. Our code is available at https://github.com/wang-zhanyu/R2GenGPT.
\end{abstract}

% 1) It significantly outperforms current state-of-the-art (SOTA) models. For instance, we outperform MSAT~\cite{wang2022medical} by 15\% in overall performance on the MIMIC-CXR dataset, with an approximate 31\% increase in the Bleu\_4 metric.


% \begin{highlights}
% \item highlight-1
% \item highlight-2
% \item highlight-3
% \end{highlights}

% Keywords

\maketitle

% Main text
\section{Introduction}
% $f({\mathbf x}_i) = {\mathbf w}^{\top}{\mathbf x}_i+ {\mathbf w}_0$\\
The landscape of radiological imaging data is experiencing exponential growth that far surpasses the availability of trained readers, resulting in a significant and unsustainable surge in radiologists' workloads. This surge in both the volume and complexity of cases places big pressure on radiologists to interpret more studies within increasingly tight timeframes. Consequently, radiologists are faced with extended working hours and a heightened risk of reading fatigue, all of which significantly contribute to diagnostic errors. Notably, the situation is particularly precarious during on-call hours for emergency radiology studies. As a result, the demand for automated radiographic report generation has soared, as it promises to alleviate the burden on radiologists, mitigate diagnostic errors, and expedite the clinical workflow.
Automated radiographic report generation (R2Gen) is a complex AI task. It aims to produce a coherent paragraph that captures the observations and findings depicted in a given radiology image. There are different R2Gen approaches based on whether the report generation is structured and whether it is template-based. This paper focuses on unstructured multi-sentence report generation.

%This technology holds significant potential in alleviating the workload and reducing diagnostic errors for radiologists who often face the challenging task of interpreting intricate studies within tight timeframes, particularly in emergency radiology reporting.

% Given its crucial clinical relevance, the domain of medical report generation has garnered increasing attention. The encoder-decoder paradigm has taken the forefront in this filed, drawing inspiration from its success in general image captioning~\cite{2015DeepVisualSemantic,vinyals2015showandtell,xu2016showattandtell,2016semanticattention}. Numerous approaches adopt analogous architectures, incorporating an encoder built upon convolutional neural networks (CNNs) and a decoder rooted in recurrent neural networks (RNNs)\cite{2015DeepVisualSemantic,vinyals2015showandtell,xu2016showattandtell}. Moreover, several research endeavors integrate meticulously designed attention mechanisms to enhance performance\cite{2017Knowing,topdown,2017Ontheautomatic}. More recently, exploratory efforts have emerged in embracing a transformer-based framework~\cite{2020Meshed,chen2020generating,CVPR201_PPKD} within this domain.

Given its critical clinical relevance, the field of medical report generation has been garnering increasing attention. Most methodologies are inspired by image/video captioning and adopt the encoder-decoder paradigm~\cite{2015DeepVisualSemantic,vinyals2015showandtell, xu2016showattandtell,2016semanticattention, tang2021clip4caption, tang2022stay}, with specific improvements tailored to the unique characteristics of the R2Gen task. In summary, recent works in the R2Gen task mainly aim to tackle two major challenges. The first challenge lies in \textbf{long text generation}. Unlike the image captioning task which generates a single sentence description, medical report generation requires detailed and coherent paragraph-long descriptions. This requires the model to have a robust capacity for learning long-range dependencies. To address this, many solutions have been proposed~\cite{2017Ontheautomatic,2020icdmAutomatic, Wang_2021_CVPR,chen2020generating,chen2022cross}. For instance, some research works~\cite{2017Ontheautomatic,2020icdmAutomatic, Wang_2021_CVPR} have employed hierarchically structured LSTM which first produces topic vectors using a sentence LSTM and then creates a description for each generated topic with a word LSTM. Another type of work R2Gen~\cite{chen2020generating} introduced a memory-driven Transformer that can record key information of the generation process, enhancing the model's ability to produce long texts. The second challenge lies in the \textbf{bias in visual and textual data}.  Due to an over-representation of normal samples in the training data, the model's learning process was biased towards these samples, limiting its ability to effectively detect abnormalities and anomalies within the dataset. Some works~\cite{Wang_2021_CVPR, 2022AlignTransformer, wang2022tmi} have addressed this issue by aligning image and text/report features, such as work Self-boost~\cite{Wang_2021_CVPR} incorporating an image-text matching branch to enhance the model's capability to capture the anomalous features in the image. Other research works mitigate the effects of data bias by incorporating external knowledge, such as medical tags~\cite{2017Ontheautomatic, wang2022medical}, and knowledge graphs~\cite{li2019knowledgedriven, 2020When, CVPR201_PPKD, 2021Knowledge}. For instance, PPKED~\cite{CVPR201_PPKD} utilizes a knowledge graph and introduces the Posterior-and-Prior Knowledge Exploring-and-Distilling framework. Despite many efforts and solutions putting forth, the aforementioned two challenges remain significant issues in this field.

Recently, large language models (LLMs) (e.g., \cite{chowdhery2022palm, touvron2023llama}) have demonstrated excellent capabilities to perform tasks with zero in-domain data, conduct logical reasoning, and apply commonsense knowledge in NLP tasks~\cite{kojima2022large, wei2022emergent}. This leads us to ponder whether we can apply large language models to medical report generation tasks, as pre-trained large language models seem to inherently possess the ability to address the two challenges mentioned above. As for long text generation, LLMs are equipped with an inherent understanding of grammar, syntax, and semantic coherence, making them well-suited for tasks requiring extended text generation, such as medical reporting. Furthermore, their proficiency in context modeling allows them to maintain consistency and relevance throughout a lengthy report. As for the bias stemming from an over-representation of normal samples in medical datasets, LLMs can serve as potential correctives due to their extensive knowledge base. Having been exposed to vast amounts of data, LLMs demonstrate robustness and are less susceptible to the effects of imbalanced datasets. They are even capable of handling numerous zero-shot tasks. Moreover, current methods mitigating bias entail the incorporation of external knowledge, whereas pre-trained LLMs inherently possess a wealth of informative knowledge.

However, applying LLMs to R2Gen tasks poses challenges due to the fundamental disparity between visual and textual modalities. The crucial step in applying LLMs to R2Gen is to bridge the gap between visual information and textual generation.  In this paper, we present R2GenGPT and explore three methods for aligning visual features with large language models. We first
process chest x-ray images using a Visual Encoder to obtain visual embeddings. These embeddings are then mapped to the LLM’s feature space via a Visual Mapper, ensuring uniform dimensions. To identify the most efficient method of aligning visual features with the LLM, we've crafted three alignment modules: 1) shallow alignment, where only the Visual Mapper is trained and other parameters remain fixed; 2) deep alignment, where both the visual encoder and the Visual Mapper are trained simultaneously; and 3) Delta alignment, where the Visual Mapper and a limited set of incremental parameters from the visual encoder are trained, ensuring both effectiveness and efficiency.

Our main contributions are summarized as follows.

$\bullet$  We propose a novel LLMs-based Radiology report generation (R2Gen) framework, dubbed R2GenGPT. This marks the first instance of harnessing pre-trained large language models (LLMs) for the R2Gen task with comprehensive comparisons conducted on two frequently employed benchmark datasets.

$\bullet$  We explored three methods with varying levels of trainable parameters to connect image modalities to large language models, namely: shallow alignment, delta alignment, and deep alignment, enabling the LLM to effectively process visual information. 
% Remarkably, our findings indicate that the delta alignment method saves 95\% of trainable parameters compared to deep alignment, achieving competitive performance with the state-of-the-art (SOTA) using only 5M trainable parameters.

$\bullet$ Our approach exhibits promising and robust performance on two widely recognized benchmark datasets—IU-Xray and MIMIC-CXR. In comparison to multiple state-of-the-art methods, our framework consistently demonstrates its efficacy, affirming its potential in the field of R2Gen.
%Our approach shows promising performance on two widely used benchmarks IU-Xray and MIMIC-CXR over multiple state-of-the-art methods.

\section{Relate Works}

\noindent \textbf{Radiology report generation} 
Radiology report generation (R2Gen) has gained significant attention in recent years, with many models being developed based on the encoder-decoder architecture initially used for image captioning tasks~\cite{vinyals2015showandtell, xu2016showattandtell, pan2020xlinear}. However, R2Gen poses additional challenges compared to image captioning, as medical reports are typically longer and clinical abnormalities in medical images are harder to detect than natural objects due to the data bias existed in the training set. To address these challenges, researchers have proposed various methods.

In~\cite{Wang_2021_CVPR}, Wang et al. introduced an image-text matching branch to facilitate report generation, utilizing report features to augment image characteristics and consequently minimize the impact of data bias. They also employed a hierarchical LSTM structure for the generation of long-form text. 
Chen et al.~\cite{chen2020generating} and Wang et al.~\cite{wang2022medical}  introduced additional memory modules to store past information, which can be utilized during the decoding process to improve long-text generation performance. 

Another type of work aims to mitigate data bias by incorporating external knowledge information, with the most representative approach being the integration of knowledge graphs~\cite{li2019knowledgedriven, 2020When, CVPR201_PPKD, 2021Knowledge, li2023dynamic, huang2023kiut}. Zhang et al.~\cite{2020When} and Liu et al.~\cite{CVPR201_PPKD} combined pre-constructed graphs representing relationships between diseases and organs using graph neural networks, enabling more effective feature learning for abnormalities. Li et al.~\cite{li2023dynamic} developed a dynamic approach that updates the graph with new knowledge in real-time. Huang et al.~\cite{huang2023kiut} incorporated knowledge from a symptom graph into the decoding stage using an injected knowledge distiller.

Apart from knowledge graphs, another method for integrating external knowledge involves incorporating semantic information to assist report generation through multi-task learning, such as multi-label classification~\cite{2017Ontheautomatic, wang2022medical,wang2022tmi, jin2023promptmrg, tanida2023interactive}. Wang et.al.~\cite{wang2022medical} extracted 768 high-frequency medical terms from RadGraph~\cite{jain2021radgraph} and trained a multi-label classification network. The prediction results from classification were then incorporated as semantic information input to the decoder, assisting the generation of the report. Jin et al.~\cite{jin2023promptmrg} make use of the diagnostic results from the classification via prompts to explicitly guide the generation process. All the above methods employing an encoder-decoder architecture in a traditional way, typically assign equal importance to both the encoder and the decoder, with a comparable number of trainable parameters. In these methods, the output from the encoder serves as the key and value for cross-attention computation in the decoder. In contrast, our approach based on LLM deviates significantly from this traditional encoder-decoder framework. Firstly, the number of parameters in the decoder significantly exceeds that in the encoder. Secondly, the encoder functions more like a ``visual tokenizer", converting images into visual tokens that are fed into LLM. The attention mechanism employed within this framework remains self-attention rather than cross-attention. With this innovative approach, our paper pioneers the use of a ``decoder-centric" architecture for the task of medical report generation.

% All the above methods adopt an encoder-decoder architecture. In this paper, we utilize the Frozen Large Language Model (LLM) to pioneer the use of a decoder-centric architecture for the task of medical report generation.\\

% The generation of medical reports has greatly benefited from various methodologies. Many are based on the hierarchical structured LSTM network, with Jing et al. and Yin et al. proposing multi-task models and topic matching mechanisms, respectively~\cite{2017Ontheautomatic,2020icdmAutomatic}. Xue et al. suggested a different network structure involving a sentence generative model and a recurrent paragraph generative model~\cite{2018Multimodal, 2019ipmiImproved}. Some studies improved performance by using transformers instead of LSTM as the text decoder~\cite{chen2020generating, CVPR201_PPKD, 2022AlignTransformer, chen2022cross, wang2022tmi, wang2023metransformer}. Recent works have integrated knowledge graphs into the medical report generation pipeline to improve report quality~\cite{2020When, 2021Radiology, 2021Knowledge, 2021Auto}. Examples include Yang et al.'s knowledge base updating mechanism, and Liu et al.'s unsupervised medical report generation pipeline~\cite{2021Radiology, 2021Auto}. All the above methods adopt an encoder-decoder architecture. In this paper, we utilize the Frozen Large Language Model (LLM) to pioneer the use of a decoder-only structure for the task of medical report generation.

\noindent \textbf{Large language Models} Recently, there has been a surge of interest in Large Language Models (LLMs) due to their superior efficacy in a wide array of Natural Language Processing (NLP) tasks. This began with transformer models like BERT~\cite{su2019vlbert}, GPT~\cite{radford2018gpt}, and T5~\cite{raffel2020T5}, each designed with distinct pre-training objectives. The introduction of GPT-3~\cite{brown2020gpt3} marked a significant shift, demonstrating the model's impressive zero-shot generalization capabilities, owed to a scaled-up parameter and data volume, which allowed it to excel in tasks it had not previously encountered. This catalyzed the development of several LLMs such as OPT~\cite{zhang2022opt}, BLOOM~\cite{scao2022bloom}, PaLM~\cite{chowdhery2022palm}, and LLaMA~\cite{touvron2023llama}, heralding the triumph of LLMs. In a parallel endeavor, Ouyang et al.~\cite{ouyang2022training} introduced InstructGPT, which brought human instruction and feedback into alignment with GPT-3. These advancements have been leveraged by applications like ChatGPT, which enables human-like dialogue interactions by responding to a vast spectrum of complex and nuanced questions and instructions.

\section{Methodology}

\begin{figure}[t]
\includegraphics[width=\textwidth]{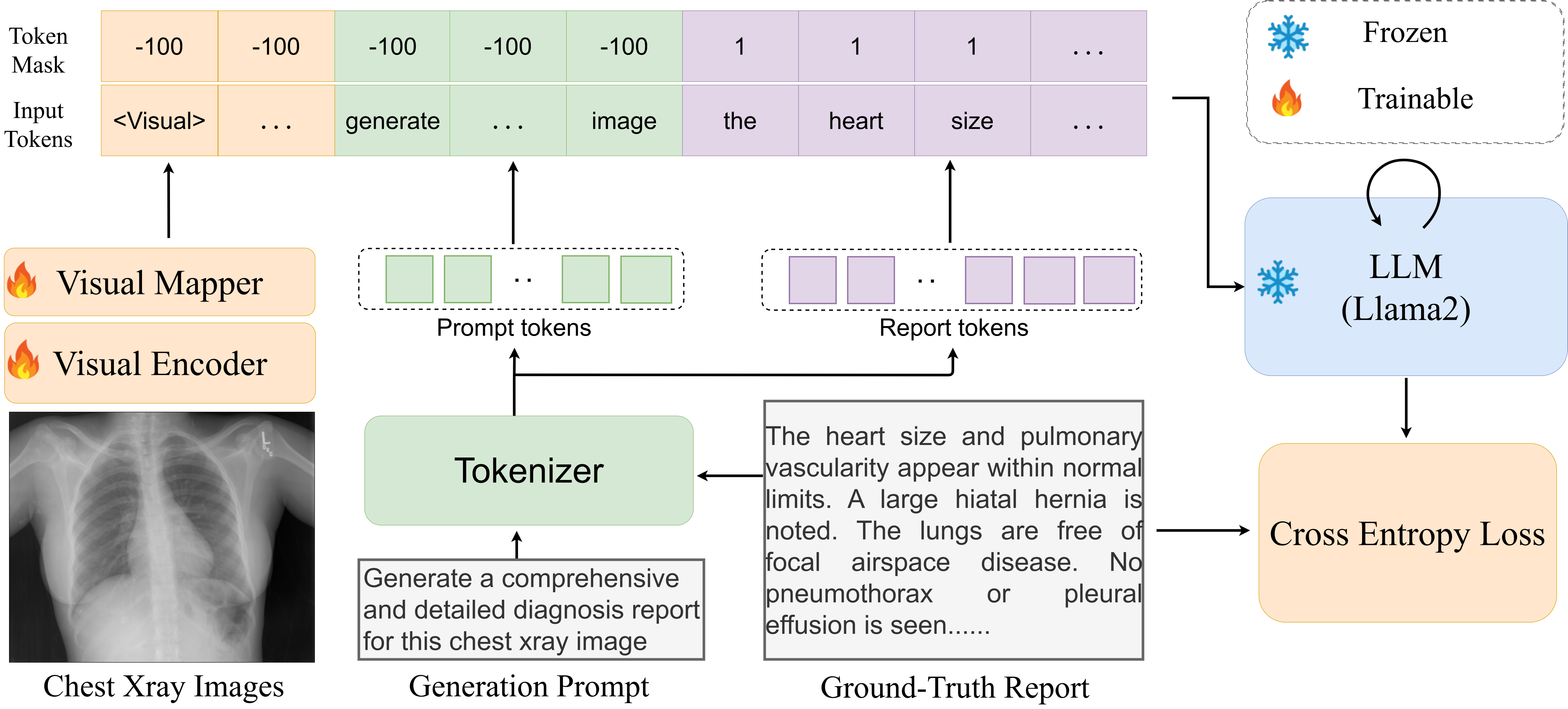}
\caption{An overview of our proposed R2GenGPT. The input tokens for the Large Language Model (LLM) are sequentially concatenated, consisting of visual tokens, prompt tokens, and report tokens. A token mask of -100 indicates that those particular tokens are excluded from auto-regressive training, while a mask of 1 signifies inclusion in auto-regressive training.}
\label{fig:framework}
\end{figure}

\noindent \textbf{Overview} As illustrated in Figure~\ref{fig:framework}, R2GenGPT comprises a Visual Encoder, a Visual Mapper, and an LLM (Large Language Model) component. The visual encoder is employed to extract information from chest x-ray images, while the visual mapper serves to project low-dimensional image features into the high-dimensional feature space of the LLM. Utilizing the visual features derived from the chest x-ray images, the LLM generates corresponding diagnostic reports.\\

\noindent \textbf{Feature Alignment} For an input chest xray image $\mathbf{X}_v$, we consider the pre-trained Swin Transformer~\cite{liu2021swin} as visual encoder, which provides the visual feature $\mathbf{Z}_v = g(\mathbf{X}_v; \theta_v)$, where $\theta_v$ is the parameters of the Swin Transformer. The grid features of the last transformer layer is utilized in our experiments. We consider a simple linear layer as the Visual Mapper to connect image features into the LLM's word embedding space. Specifically, we apply a trainable projection matrix $\mathbf{W}_m$ to convert $\mathbf{Z}_v$ into language embedding tokens $\mathbf{H}_v$, which have the same dimensionality of the word embedding space in the large language model.

\begin{equation}~\label{equ:mapper}
   \qquad\qquad\qquad\qquad\qquad\qquad\qquad \mathbf{H}_v = \mathbf{W}_m \mathbf{Z}_v, \quad \mathrm{with} \; \mathbf{Z}_v = g(\mathbf{X}_v)
\end{equation}
Thus we have a sequence of visual tokens $\mathbf{H}_v$. Following the extraction of visual tokens $\mathbf{H}_v$, we propose the following three distinct training strategies to identify the most efficient aligning method by varying the level of trainable parameters.

a) Shallow Alignment: In this mode, we fix the parameters of the pre-trained Swin Transformer and train only the linear Visual Mapper, represented by $\mathbf{W}_m$.

b) Deep Alignment: For this approach, both the Swin Transformer and the Visual Mapper are jointly fine-tuned. Specifically, parameters from both the Visual Encoder (Swin Transformer) and the Visual Mapper, denoted as $\theta_v$ and $\mathbf{W}_m$ respectively, are updated. 

c) Delta Alignment: As the Swin Transformer utilized in this paper was originally trained on natural images, the shallow alignment approach hinders the model's ability to capture high-quality radiographic image features. On the other hand, adopting deep alignment substantially impacts the model's training efficiency. Therefore, we propose delta alignment, parameter-efficiently fine-tuning the Swin Transformer model using LoRA~\cite{hu2022lora}. Specifically, for a pre-trained weight matrix $\mathbf{W}_0$ within $\theta_v$, LoRA constrains its update with two smaller matrices using a low-rank decomposition $\mathbf{W}_0 + \Delta\mathbf{W}_0 = \mathbf{W}_0 + \mathbf{BA}$, where $\mathbf{W}_0 \in \mathbb{R}^{d \times k}$, $\mathbf{B} \in \mathbb{R}^{d \times r}$, $\mathbf{A} \in \mathbb{R}^{r \times k}$, and the rank $r \ll min(d, k)$. It is noted that in our implementation, we only adjust the query and value projections within the swin transformer to prioritize a simple yet efficient model. The trained parameters are denoted as $\Delta\theta_v$, and both $\Delta\theta_v$ and $\mathbf{W}_m$ are trained in this mode.\\

%------------------------------------------------------------------

\noindent \textbf{Large Language Models} We adopt Llama2-7B model for the large language model component. The Llama2-7B stands out for its remarkable capabilities and robustness. Designed with a massive 7-billion-parameter architecture, it encapsulates a rich knowledge base derived from extensive pre-training on diverse datasets. One of its key strengths lies in its extraordinary ability to understand and generate complex language structures, making it particularly well-suited for intricate tasks such as radiology report generation. 

% The model's fine-grained attention mechanisms, coupled with its scalable design, allow for nuanced comprehension of both textual and visual information. Moreover, the llama2-7B's high efficiency in training, as well as its adaptability to various domains, accentuates its appeal as a choice in our system. By integrating llama2-7B, R2GenGPT is able to leverage state-of-the-art natural language processing and computer vision techniques, achieving a synthesis of visual understanding and linguistic description that pushes the boundaries of medical image report generation.

Given an chest xray image $\mathbf{X}_v$ and its corresponding report $\mathbf{X}_r$, the detailed prompt inputted into Llama2 is as follows.

\begin{center}
    \textit{Human: <Img>$\mathbf{X}_v$</Img>, $\mathbf{X}_p$.  \textbackslash n Assistant: $\mathbf{X}_r$ </s>.}
\end{center}

Here $\mathbf{X}_p$ is our designed instruction prompt specific to the R2Gen task. In our current implementation, $\mathbf{X}_p$ = "Generate a comprehensive and detailed diagnosis report for this chest xray image.". For this prompt, before inputting it into LLAMA2 for computation, $\mathbf{X}_v$ will be replaced by visual tokens $\mathbf{H}_v$ processed using Equ.~\ref{equ:mapper} while all other text is tokenized into word tokens using LLAMA's tokenizer.\\

\noindent \textbf{Loss Function}~~~We perform instruction-tuning of the LLM only on the report tokens , using its original auto-regressive training objective. Specifically, for a report of length $L$, conditioned on visual information $\mathbf{X}_v$ and instruction prompt $\mathbf{X}_p$, our loss function, captured as the negative log likelihood, is formulated as:
\begin{equation}~\label{equ:loss}
\qquad\qquad\qquad\qquad\qquad\qquad \mathcal{L}(\theta; \mathbf{X}_r, \mathbf{X}_v, \mathbf{X}_p) = -\sum_{i=1}^{L} \log p_\theta (x_i|\mathbf{X}_v,\mathbf{X}_p,\mathbf{X}_{r,<i}),
\end{equation}
% \begin{equation} 
%     \qquad\qquad\qquad\qquad\qquad\qquad\qquad p(X_r|X_v, X_p) = \prod_{i=1}^{L}p_\theta (x_i|X_v,X_p,X_{r,<i}) 
% \end{equation} 
where $\theta$ is the trainable parameters, $\mathbf{X}_{r, <i}$ is the report tokens before the current prediction token $x_i$.

%------------------------------------------------------------------
\section{Experiments}~\label{Sec:Experiments}
\subsection{Data Collection}~\label{sec:3-1}
We evaluated performance using two datasets: a widely-used benchmark IU-Xray~\cite{2015iu-xray} and the currently largest dataset MIMIC-CXR~\cite{2019MIMIC} for medical report generation.\\

\noindent\textbf{IU-Xray}:~~~Indiana University Chest X-ray Collection (IU-Xray) \cite{2015iu-xray} is the most widely used publicly accessible dataset in medical report generation tasks. It contains 3,955 fully de-identified radiology reports, each of which is associated with frontal and/or lateral chest X-ray images, and 7,470 chest X-ray images in total. Each report is comprised of several sections: Impression, Findings, Indication, etc. In this work, we adopt the same data set partitioning as~\cite{chen2020generating} for a fair comparison, with a train/test/val set by 7:1:2 of the entire dataset.  All evaluations are done on the test set.\\

\noindent\textbf{MIMIC-CXR}:~~~This largest publicly available dataset encompasses both chest radiographs and unstructured textual reports. This comprehensive dataset comprises a total of 377,110 chest X-ray images and 227,835 corresponding reports sourced from 64,588 patients who underwent examination at the Beth Israel Deaconess Medical Center between 2011 and 2016. To ensure equitable comparisons, we adhered to MIMIC-CXR's official partitioning as outlined in~\cite{chen2020generating}, resulting in 270790 samples designated for training, while allocating 2130 and 3,858 samples for validation and testing, respectively.

\begin{table}
\centering
\caption{Comparison on IU-Xray (upper part) and MIMIC-CXR datasets (lower part). $\dagger$ indicates the results are quoted from their respective papers. For the methods without $\dagger$, their results are obtained by re-running the publicly released codebase~\cite{li2021codebase} on these two datasets using the same training-test partition as our method. }
\label{Table:ComparisonWithSOTA}
\begin{tblr}{
  cell{1-27}{3-9} = {c},
  vline{2-3} = {-}{},
  hline{1-2,14,28} = {-}{},
  hline{11,25} = {2-9}{},
}
Dataset   & Methods             & BLEU-1         & BLEU-2         & BLEU-3         & BLEU-4         & ROUGE          & METEOR         & CIDEr          \\
          & Show-Tell~          & 0.243          & 0.130          & 0.108          & 0.078          & 0.307          & 0.157          & 0.197          \\
          & Att2in~             & 0.248          & 0.134          & 0.116          & 0.091          & 0.309          & 0.162          & 0.215          \\
          & AdaAtt~             & 0.284          & 0.207          & 0.150          & 0.126          & 0.311          & 0.165          & 0.268          \\
          & Transformer~        & 0.372          & 0.251          & 0.147          & 0.136          & 0.317          & 0.168          & 0.310          \\
          & M2transformer       & 0.402          & 0.284          & 0.168          & 0.143          & 0.328          & 0.170          & 0.332          \\
IU-Xray   & R2Gen$^\dagger$~    & 0.470          & 0.304          & 0.219          & 0.165          & 0.371          & 0.187          & -              \\
          & R2GenCMN$^\dagger$~ & 0.475          & 0.309          & 0.222          & 0.170          & 0.375          & 0.191          & -              \\
          & MSAT$^\dagger$~               & 0.481          & 0.316          & 0.226          & 0.171          & 0.372          & 0.190          & 0.394          \\
          & METransformer$^\dagger$       & 0.483          & \textbf{0.322} & 0.228          & 0.172          & \textbf{0.380} & 0.192          & 0.435          \\
          & R2GenGPT (Shallow)  & 0.466          & 0.301          & 0.211          & 0.156          & 0.370          & 0.202          &  0.405         \\
          & R2GenGPT (Delta)    & 0.470          & 0.299          & 0.213          & 0.162          & 0.369          & 0.211          & 0.419          \\
          & R2GenGPT (Deep)     & \textbf{0.488} & 0.316          & \textbf{0.228} & \textbf{0.173} & 0.377          & \textbf{0.211} & 0.\textbf{438} \\
          & Show-Tell~          & 0.308          & 0.190          & 0.125          & 0.088          & 0.256          & 0.122          & 0.096          \\
          & Att2in~             & 0.314          & 0.198          & 0.133          & 0.095          & 0.264          & 0.122          & 0.106          \\
          & AdaAtt~             & 0.314          & 0.198          & 0.132          & 0.094          & 0.267          & 0.128          & 0.131          \\
          & Transformer~        & 0.316          & 0.199          & 0.140          & 0.092          & 0.267          & 0.129          & 0.134          \\
          & M2Transformer~      & 0.332          & 0.210          & 0.142          & 0.101          & 0.264          & 0.134          & 0.142          \\
          & R2Gen$^\dagger$~    & 0.353          & 0.218          & 0.145          & 0.103          & 0.277          & 0.142          & -              \\
MIMIC-CXR & R2GenCMN$^\dagger$~ & 0.353          & 0.218          & 0.148          & 0.106          & 0.278          & 0.142          & -              \\
          & PPKED$^\dagger$~    & 0.36           & 0.224          & 0.149          & 0.106          & 0.284          & 0.149          & 0.237          \\
          & GSK$^\dagger$~      & 0.363          & 0.228          & 0.156          & 0.115          & 0.284          & -              & 0.203          \\
          & MSAT$^\dagger$~     & 0.373          & 0.235          & 0.162          & 0.120          & 0.282          & 0.143          & 0.299          \\
          & METransformer$^\dagger$       & 0.386          & 0.250          & 0.169          & 0.124          & 0.291          & 0.152          & \textbf{0.362}          \\
          & R2GenGPT (Shallow)  & 0.365          &  0.237         & 0.163          & 0.117          &  0.277         & 0.136           & 0.145               \\
          & R2GenGPT (Delta)    & 0.380          & 0.244          & 0.167          & 0.119          &  0.281         & 0.145           &  0.195              \\
          & R2GenGPT (Deep)     & \textbf{0.411} & \textbf{0.267} & \textbf{0.186} & \textbf{0.134} & \textbf{0.297} & \textbf{0.160}  &  0.269              
\end{tblr}
\end{table}

% -----------------------------------------------------------------
\begin{figure}[t]
\includegraphics[width=\textwidth]{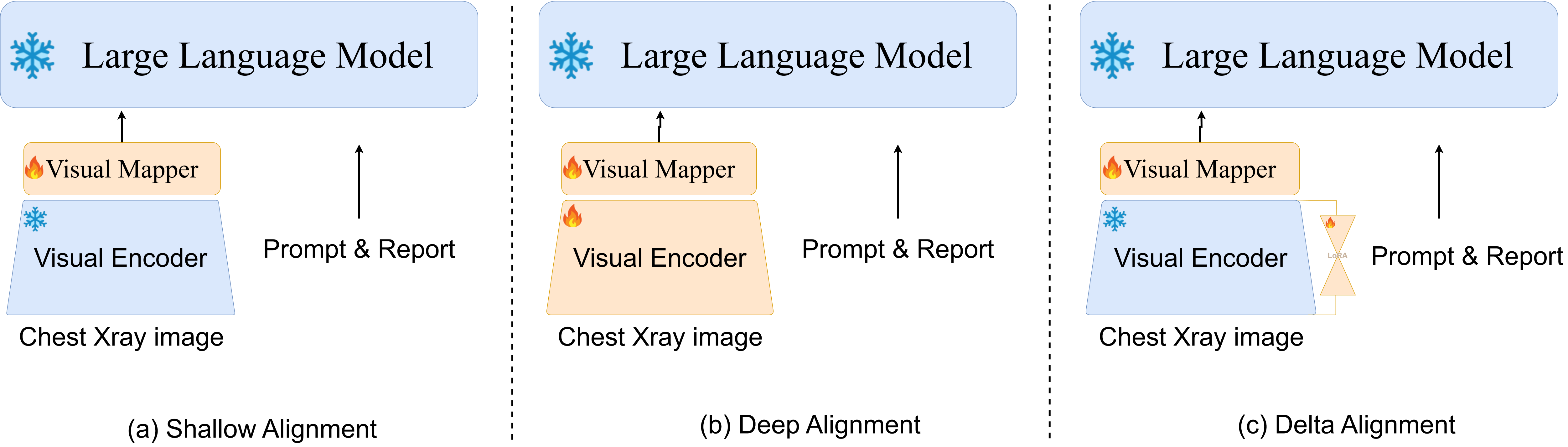}
\caption{Three proposed alignment methods. (a) Shallow Alignment: Training only the Linear Layer. (b) Deep Alignment: Training both the Linear Layer and all parameters of the Visual Encoder. (c) Delta Alignment: Training the Linear Layer and a small subset of incremental parameters of the Visual Encoder.}
\label{fig:align}
\end{figure}

\subsection{Experimental Settings}
\noindent \textbf{Evaluation Metrics}~~~Adhering to the established evaluation protocol~\footnote{https://github.com/tylin/coco-caption}, we employ the prevalent metrics for assessment, namely BLEU scores~\cite{Kishore2002bleu}, ROUGE-L~\cite{lin-2004-rouge}, METEOR~\cite{banerjee-lavie-2005-meteor} and CIDEr~\cite{vedantam2015cider}, to gauge the quality of the generated textual reports. To measure the accuracy of descriptions for clinical abnormalities, we follow~\cite{chen2020generating, chen2022cross, 2021Auto} and further report clinical efficacy metrics. Specifically, we employ CheXpert~\cite{irvin2019chexpert} for annotating the generated reports, which are subsequently compared against ground truth annotations across 14 distinct categories related to thoracic diseases and support devices. We use precision, recall, and F1 to evaluate model performance for clinical efficacy metrics.

\noindent \textbf{Implementation Details}~~~In this work, we leveraged the LLAMA2-7B model~\footnote{https://huggingface.co/meta-llama/Llama-2-7b-chat-hf} as the large language model and the base version of the Swin Transformer~\footnote{https://huggingface.co/microsoft/swin-base-patch4-window7-224} as the Visual Encoder. Within the parameters of LoRA, we configured the Lora attention dimension to 16, and the alpha parameter for Lora scaling was also set at 16. The training process was conducted on four NVIDIA A100 40GB GPUs using mixed precision for 3 epochs for MIMIC-CXR and 15 epochs for IU-Xray dataset, with a mini-batch size of 6 and a learning rate of 1e-4. During the testing phase, we employed a beam search strategy with a beam size set to 3.

\begin{figure}[t]
\includegraphics[width=\textwidth]{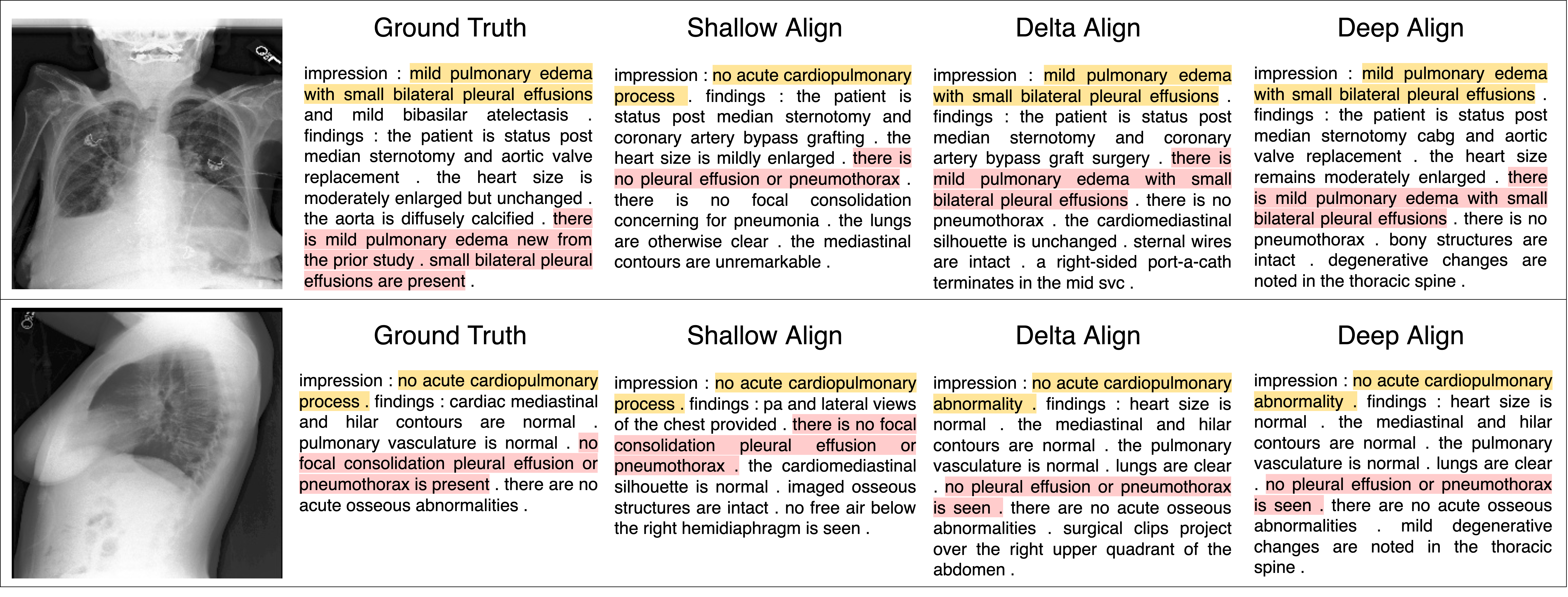}
\caption{Examples of the generated report on MIMIC-CXR dataset. We compared the results generated by the three alignment methods proposed in R2GenGPT. For better illustration, the key medical information in the reports are highlighted using different colors.}
\label{fig:examples}
\end{figure}

\subsection{Results and Discussion}
\noindent\textbf{Comparison with SOTA}~~~Table~\ref{Table:ComparisonWithSOTA} showcases a performance comparison between the state-of-the-art methods and our R2GenGPT model variants on the IU-Xray and MIMIC-CXR dataset. In terms of standard image captioning methods, the table considers Show-Tell~\cite{vinyals2015showandtell}, Att2in~\cite{xu2016showattandtell}, AdaAtt~\cite{2017Knowing}, Transformer~\cite{vaswani2017attentionisallyouneed}, and M2Transformer~\cite{2020Meshed}. Furthermore, medical report generation methods such as R2Gen~\cite{chen2020generating}, R2GenCMN~\cite{chen2022cross}, MSAT~\cite{wang2022medical}, METransformer~\cite{wang2023metransformer}, and other methods marked with $\dagger$ in Table~\ref{Table:ComparisonWithSOTA} are considered. 

From Table~\ref{Table:ComparisonWithSOTA}, it is evident that our R2GenGPT model variants, especially R2GenGPT (Deep), outperform the compared methods across nearly all evaluation metrics. In the MIMIC-CXR dataset, apart from CIDEr, we significantly outperform the latest METransformer~\cite{wang2023metransformer} method across all metrics. For instance, our BLEU\_4 score is improved from 0.124 to 0.134, marking an 8.1\% increase. However, we achieved a CIDEr score of 0.269, which is lower than METransformer's 0.362. This discrepancy is because METransformer employs an expert voting strategy similar to an ensemble approach to enhance the CIDEr metric. In comparison to methods without this enhancement, such as R2Gen~\cite{chen2020generating} and PPKED~\cite{CVPR201_PPKD}, we also hold a distinct advantage in terms of the CIDEr metric. It is also noteworthy that our R2GenGPT (Shallow), with only 4.2M trainable parameters, has been able to achieve a performance in par with the well-known R2Gen model~\cite{chen2020generating}, if not even better.\\

\begin{table}
\centering
\caption{Evaluation of Model Efficiency and Clinical Efficacy on MIMIC-CXR dataset.}
\label{tab:eff}
\begin{tblr}{
  colsep = 6pt,
  cells = {c},
  cell{1}{1} = {r=2}{},
  cell{1}{2} = {c=3}{},
  cell{1}{5} = {c=2}{},
  cell{1}{7} = {c=3}{},
  vline{2,5,7} = {1-6}{},
  hline{1,3,6-7} = {-}{},
}
Models        & Trainable Components &                 &              & Scale and Efficiency &                 & Clinical Efficacy &        &       \\
              & Mapper               & Encoder         & LoRA         & Trainable Parameter  & Time~~          & Precision         & Recall & F1    \\
Shallow       & \checkmark           &                 &              & 4.2M                 & 1.75h/epo       & 0.341             & 0.312  & 0.325 \\
Delta         & \checkmark           &                 & \checkmark   & 5.0M                 & 1.83h/epo       & 0.366             & 0.350  & 0.358 \\
Deep          & \checkmark           & \checkmark      &              & 90.9M                & 2.75h/epo       & \textbf{0.392}    & \textbf{0.387}  & \textbf{0.389} \\
METransformer & -                    & -               & -            & 152M                 & 3.62h/epo       & 0.364             & 0.309  & 0.334 
\end{tblr}
\end{table}

% -----------------------------------------------------------------
\noindent\textbf{Model Efficiency and Clinical Efficacy Analysis}~~~In Table~\ref{tab:eff}, we have presented both model efficiency and clinical efficacy metrics. It's evident that our model exhibits higher training efficiency compared to METransformer. For instance, R2GenGPT (Deep) requires training with only 90.9 million parameters, which is significantly less than METransformer's 152 million. Furthermore, R2GenGPT (Delta) achieves comparable performance with just 5 million parameters. To assess the model's training efficiency, we conducted evaluations on four A100 40G GPUs and recorded the time required for one training epoch. Notably, R2GenGPT Shallow, Delta, and Deep each completed this epoch in just 1.75, 1.83, and 2.75 hours, respectively, compared to METransformer's 3.62 hours, highlighting our model's superior training efficiency. In terms of clinical efficacy metrics, it can be observed that our R2GenGPT(deep) and R2GenGPT(delta) achieved F1 scores of 0.389 and 0.358, respectively, surpassing the current SOTA method, METransformer, with a score of 0.334. This demonstrates the ability of R2GenGPT to generate crucial clinical information.\\

\noindent\textbf{Qualitative results}~~~In Figure~\ref{fig:examples}, we compare the reports generated by the three alignment methods of R2GenGPT. To provide a better visualization, we  highlight key medical information in both the ground truth and generated reports using different colors. From the figure, it can be observed that reports generated using the Shallow Alignment method are notably inferior to those generated using the Delta Alignment and Deep Alignment methods. For instance, in the first example (top), the Shallow Alignment method erroneously identifies a sample with mild pulmonary edema as a normal sample, whereas Delta Alignment and Deep Alignment methods can accurately identify it.

\section{Conclusions}
In this paper, we present R2GenGPT, an innovative framework at the forefront of Radiology Report Generation (R2Gen) that capitalizes on the capabilities of Large Language Models (LLMs). Through a comprehensive exploration of three alignment methods—shallow, delta, and deep— this research highlights the game-changing potential of LLMs in elevating the R2Gen landscape. R2GenGPT not only attains competitive SOTA performance but also achieves a remarkable reduction in computational complexity. This dual achievement positions R2GenGPT as a promising solution to automate and improve radiology reporting.
\bibliographystyle{cas-model2-names}

\bibliography{cas-refs}



\end{document}